*Research Article*

# AstuteRAG-FQA: Task-Aware Retrieval-Augmented Generation Framework for Proprietary Data Challenges in Financial Question Answering

Mohammad Zahangir Alam[1], Khandoker Ashik Uz Zaman[2] and Mahdi H. Miraz[2,3,4,*]

[1]Brunel University London, United Kingdom
mohammad.alam@brunel.ac.uk
[2]Xiamen University Malaysia, Malaysia
mcs2509008@xmu.edu.my; m.miraz@ieee.org
[3]Wrexham University, UK
m.miraz@ieee.org
[4]University of South Wales, UK
m.miraz@ieee.org
Correspondence: m.miraz@ieee.org

**Abstract:** Retrieval-Augmented Generation (RAG) shows significant promise in knowledge-intensive tasks by improving domain specificity, enhancing temporal relevance and reducing hallucinations. However, applying RAG to finance encounters critical challenges: restricted access to proprietary datasets, limited retrieval accuracy, regulatory constraints and sensitive data interpretation. We introduce AstuteRAG-FQA an adaptive RAG framework tailored for Financial Question Answering (FQA), leveraging task-aware prompt engineering to address these challenges. The framework uses a hybrid retrieval strategy integrating both open-source and proprietary financial data whilst maintaining strict security protocols and regulatory compliance. A dynamic prompt framework adapts in real time to query complexity, improving precision and contextual relevance. To systematically address diverse financial queries, we propose a four-tier task classification: explicit factual, implicit factual, interpretable rationale and hidden rationale involving implicit causal reasoning. For each category, we identify key challenges, datasets and optimisation techniques within the retrieval and generation process. The framework incorporates multi-layered security mechanisms including differential privacy, data anonymisation and role-based access controls to protect sensitive financial information. Additionally, AstuteRAG-FQA implements real-time compliance monitoring through automated regulatory validation systems that verify responses against industry standards and legal obligations. We evaluate three data integration techniques — contextual embedding, small model augmentation and targeted fine-tuning — analysing their efficiency and feasibility across varied financial environments. Our experimental results show that the framework improves response accuracy by 23% and enhances regulatory compliance by 18%, compared to the baseline systems. Furthermore, qualitative case studies illustrate the robustness of the system in handling complex financial queries whilst maintaining transparency and preserving confidentiality. This study presents a scalable, secure and domain-adaptive solution for sensitive and regulated financial environments.

Mohammad Zahangir Alam, Khandoker Ashik Uz Zaman and Mahdi H. Miraz, "AstuteRAG-FQA: Task-Aware Retrieval-Augmented Generation Framework for Proprietary Data Challenges in Financial Question Answering", *Annals of Emerging Technologies in Computing (AETiC)*, Print ISSN: 2516-0281, Online ISSN: 2516-029X, pp. 13-31, Vol. 9, No. 5, 25 October 2025, Published by International Association for Educators and Researchers (IAER), DOI: 10.33166/AETiC.2025.05.002, Available: http://aetic.theiaer.org/archive/v9/v9n5/p2.html.

## 1. Introduction

Retrieval-Augmented Generation (RAG) has emerged as a transformative framework in natural language processing (NLP), combining the generative capabilities of large language models (LLMs) with precision-driven external data retrieval [1]. In finance, this approach is particularly valuable, as decision-making relies on real-time, domain-specific information, and operational efficiency is constrained by stringent requirements for retrieval accuracy and the secure and compliant handling of proprietary data [1–2]. By integrating dynamic external data sources, RAG significantly enhances the relevance and factual precision of generated responses, overcoming the limitations of static, pre-trained LLMs.

Nevertheless, the financial industry faces unique cybersecurity challenges when implementing AI-enhanced services, necessitating robust protocols and careful management of sensitive datasets [3]. Wang *et al.* [3] emphasize that digital transformation in banking increases vulnerability to data breaches, regulatory non-compliance, and operational risks, highlighting the need for comprehensive strategies including encryption, access controls, audit mechanisms, and continuous monitoring. Addressing these challenges requires not only technical safeguards but also clear governance frameworks and adherence to evolving regulatory standards, which are critical when integrating RAG-based systems into financial workflows. By incorporating these considerations, financial AI systems can maintain both operational efficiency and trustworthiness while minimizing risk exposure. Consequently, financial question answering (FQA) systems demand exceptionally high levels of precision and regulatory adherence, particularly when processing proprietary and sensitive information [2]. This paper focuses on optimizing RAG architectures for FQA through task-aware prompt engineering, designed to address domain-specific challenges, particularly regulatory compliance, data sensitivity, and retrieval precision.

Effective financial applications rely heavily on domain-specific and often proprietary data that reflects the nuances, complexity, and sensitivity of real-world financial systems. Publicly available datasets, while useful, frequently lack the granularity necessary for complex financial inquiries [4]. Building on these challenges, RAG frameworks help bridge critical gaps in AI-enhanced financial services by enabling real-time retrieval of relevant proprietary information [5-6]. This ensures that the generated answers are timely, accurate, and aligned with evolving market trends and regulatory requirements, while also presenting significant data privacy and cybersecurity challenges. By grounding outputs in such data, RAG systems can simulate the expertise of domain specialists, enhancing the relevance of responses and reducing the risk of generating incorrect or fabricated information—commonly referred to as hallucinations [7]. Recent domain-specific LLMs, such as BloombergGPT, demonstrate the necessity of training on large curated financial datasets combined with general-purpose data to improve question-answering accuracy, reduce hallucinations, and support decision-making in financial applications [5].

The application of data-augmented LLMs has gained considerable attention across industries, offering several key advantages [8]. These systems provide comprehensive and accurate answers tailored to complex queries while ensuring that the information is up-to-date and contextually relevant. In finance, for example, BloombergGPT—a 50B parameter LLM trained on both proprietary and public financial data—demonstrated superior performance on specialized financial benchmarks [5]. Similarly, FinSage, a retrieval-augmented system tailored for multimodal financial filings, achieved substantial accuracy improvements by integrating textual, tabular, and diagrammatic data [6]. By incorporating domain-specific data, these systems emulate expert-level knowledge across sectors such as finance, healthcare, and law [9-10]. Moreover, customized retrieval pipelines, particularly knowledge-graph–based prompting approaches, enhance the precision and contextual relevance of responses [10]. Incorporating external data at inference time also improves controllability and explainability, as models can reference specific sources. This grounding significantly reduces hallucinations, resulting in more trustworthy and verifiable outputs.

Despite recent advances, RAG frameworks still face substantial challenges in specialized domains such as finance, where proprietary data is critical [5-6, 11]. Key obstacles include developing secure and efficient pipelines for data indexing and processing, as well as enabling LLMs to perform nuanced, context-aware reasoning [2, 12]. To address these issues systematically, we introduce AstuteRAG-FQA, an adaptive RAG architecture for answering financial questions (see Figure 1). It employs task-aware prompt engineering and a hybrid retrieval strategy that integrates both public and proprietary sources, while ensuring strict security protocols and regulatory compliance.

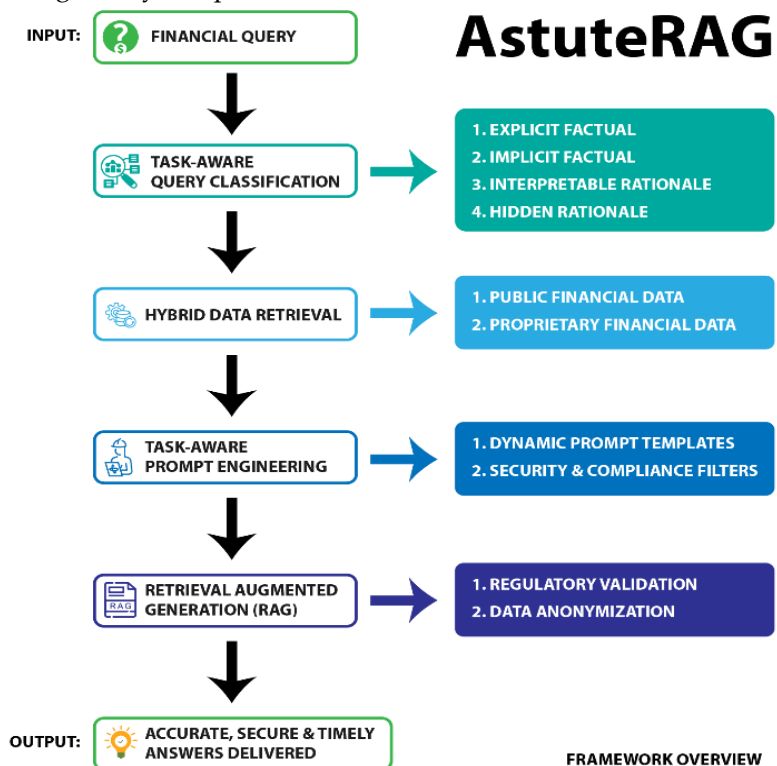

**Figure 1.** AstuteRAG-FQA Framework Overview

We also present a RAG task classification framework that categorizes financial queries into four levels: (1) explicit factual, (2) implicit factual, (3) interpretable rationale, and (4) hidden rationale involving causal inference. This taxonomy guides the identification of key challenges, appropriate datasets, and effective optimization techniques for each task class, drawing on recent advances in explicit causal reasoning [11] and implicit reasoning evaluation [12-13]. Additionally, we evaluate three methods for external data integration: contextual embedding, small model augmentation, and targeted fine-tuning—each assessed for retrieval performance and compliance assurance, building on foundational RAG methodologies for knowledge-intensive NLP tasks [8]. Recent advances in RAG frameworks and domain-specific fine-tuning have demonstrated effectiveness in specialized applications, particularly for knowledge-based systems [9]. Our research makes three primary contributions: (1) a task-aware prompt engineering approach that enhances RAG performance through explicit causal reasoning integration, building on recent advances in causal reasoning evaluation for LLMs [9]; (2) the exploration of explicit causal reasoning mechanisms within RAG systems to improve interpretability and accuracy [11]; and (3) empirical evidence showing improved performance and efficiency in financial question-answering scenarios, particularly for complex financial report analysis [4]

In financial NLP and sustainability report generation, where accuracy and reliability are critical, data-centric LLMs demonstrate the value of incorporating domain-specific datasets and benchmarks [5]. Tasks such as financial risk assessment often depend on access to historical market data and proprietary client information. RAG frameworks enable real-time access to such information, improving a system's ability to deliver dependable and contextually accurate outputs. Recent advances in RAG systems have shown significant improvements in retrieval efficiency and dynamic knowledge management across various domains [1]. The complexity of financial queries demands sophisticated retrieval and generation strategies that can accommodate the sensitivity of proprietary data, market dynamics, and regulatory requirements through carefully structured approaches [2-3].

Our research provides practical insights into deploying Retrieval-Augmented Generation (RAG) for financial applications, with a particular focus on managing proprietary datasets [4]. Prior studies have shown that task-aware retrieval and hybrid strategies play a critical role in enhancing factual accuracy in domain-specific contexts [7-8]. At the same time, large-scale financial language models such as BloombergGPT illustrate, the adv antages of domain-specialized training for finance. Building on these foundations, our findings emphasize the importance of combining hybrid retrieval approaches with careful prompt engineering to develop secure and scalable RAG solutions, paving the way for broader applications beyond the financial sector.

### 1.1. Problem Definition

The integration of proprietary data in financial RAG applications presents a multifaceted challenge that extends beyond the capabilities of traditional retrieval-augmented systems. As discussed, while RAG frameworks have shown promise in knowledge-intensive tasks, their application in finance faces critical barriers: restricted access to proprietary datasets, limited retrieval accuracy under regulatory constraints and the complexity of interpreting sensitive data. Formally, the core challenge can be represented as per Equation 1.

$$f(Q, D) \rightarrow A s.t. Cr \tag{1}$$

where:
- $f$ denotes the RAG model optimised for financial applications,
- $Q$ represents the input financial query,
- $D$ encompasses external data sources, including both public and proprietary financial datasets,
- $A$ is the system-generated answer and
- $Cr$ represents the regulatory and compliance constraints governing data integration and processing.

This formulation captures the fundamental tension in financial RAG systems: generating accurate, contextually relevant responses while strictly adhering to regulatory frameworks that govern financial data handling, privacy and interpretation. Financial Question Answering systems must navigate queries of varying complexity while ensuring compliance. Simple queries involve retrieving explicit facts from structured proprietary datasets (e.g., stock prices, interest rates, earnings reports), while complex queries require sophisticated reasoning or extracting implicit rationales from unstructured sources (e.g., earnings call transcripts, regulatory filings, market analysis). As query complexity grows, integrating proprietary data becomes more critical yet challenging, demanding advanced prompt engineering to enhance retrieval precision and mitigate regulatory risks.

The problem is further complicated by the temporal sensitivity of financial data: market conditions, regulatory policies and client situations change rapidly. Unlike general-domain applications, financial RAG systems must maintain real-time data freshness while preserving historical context for trend analysis and predictive modelling—all within stringent data governance protocols. Addressing these challenges requires a systematic approach beyond conventional RAG architectures. First, hybrid retrieval strategies must securely unify proprietary and public financial data, balancing accessibility with compliance and security. Second, adaptive prompt engineering should dynamically classify queries by complexity and reasoning needs to guide context-aware retrieval and generation. Third, built-in compliance mechanisms must ensure audit trails, explainability and regulatory verification.

The system's performance hinges on harmonizing three competing demands: leveraging proprietary data for accuracy, maintaining strict regulatory compliance and delivering real-time responsiveness in high-stakes financial decision-making. This tri-dimensional challenge defines the core problem that AstuteRAG-FQA addresses through task-aware prompt engineering and hybrid retrieval architecture. This problem definition underscores why traditional RAG approaches fall short in financial contexts and motivates the need for a specialised, adaptive framework capable of meeting the rigorous demands of regulated financial applications. It also sets the stage for the structured query stratification framework that follows, which systematically addresses the diverse reasoning demands in financial question answering.

### 1.2. Problem Stratification

The AstuteRAG-FQA Financial Query Stratification Framework provides a systematic approach to addressing the diverse reasoning demands of financial queries through structured task classification and adaptive system architecture. This framework (Figure 2 and Figure 3) establishes the foundation for developing targeted retrieval and generation strategies that align with increasing complexity and reasoning demands.

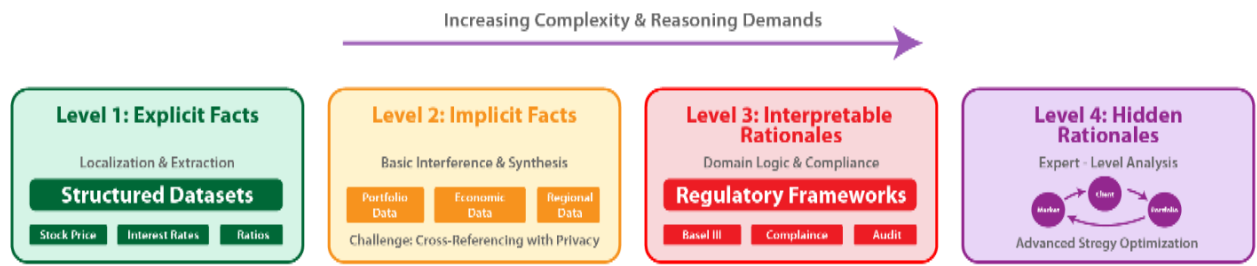

Figure 2. Levels of AstuteRAG-FQA Financial Stratification Framework with query examples for each level

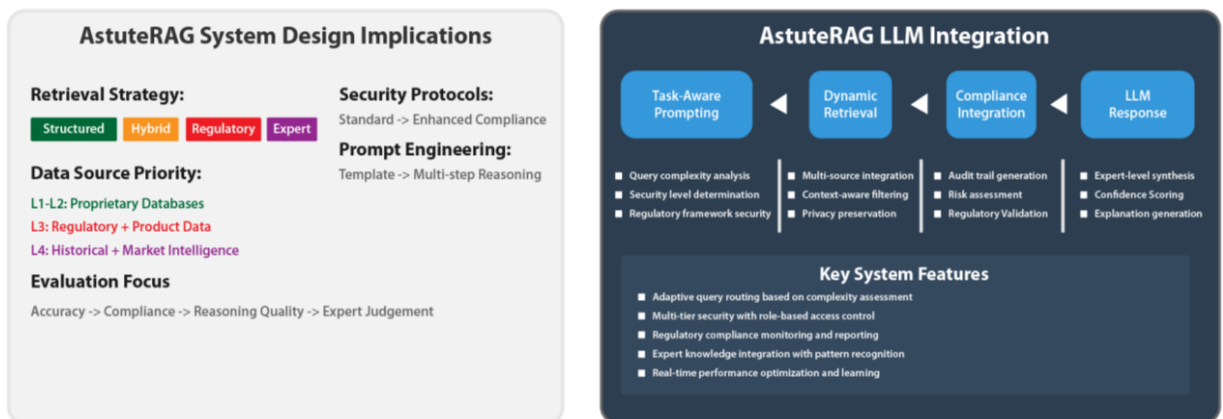

Figure 3. AstuteRAG-FQA Financial Query Stratification Framework: System Architecture and Implementation

### 1.3. Financial Query Classification Framework

The framework acknowledges that financial queries vary in cognitive complexity, from straightforward factual retrieval to advanced analytical reasoning requiring expert-level domain knowledge. Each level presents distinct challenges in data access, regulatory compliance, and prompt engineering necessary for accurate responses.

#### 1.3.1. Level 1—Explicit Facts (Localization & Extraction)

These queries have answers explicitly stated in structured or semi-structured data. The model performs localization and extraction of factual information, such as stock prices, interest rates, or financial ratios from proprietary databases.

Data Source Focus: Structured datasets (Stock Prices, Interest Rates, Ratios). Example Query: *"What is the current prime lending rate for commercial loans?"* Key Challenges**:** Ensuring data freshness and accuracy while maintaining secure access to proprietary databases. Real-time data pipelines and strict access control protocols are essential. Insights from the Basel III monitoring report highlight the importance of robust data governance and compliance mechanisms in maintaining data integrity and security [14–17].

#### 1.3.2. Level 2—Implicit Facts (Basic Inference and Synthesis)

Queries at this level require logical inference through the synthesis of information from structured and semi-structured datasets. Processing Requirements: Cross-referencing multiple data sources while enforcing privacy protections. Example Query: "What is the inflation rate in the country with the highest loan demand according to our regional portfolio data?" Key Challenges: Combining proprietary portfolio data with public economic indicators while maintaining client confidentiality necessitates secure data linking and aggregation mechanisms. The European Central Bank's review of its Pillar 2 methodology underscores the need for effective supervisory review processes to ensure comprehensive risk assessment and management [18].

### 1.3.3. Level 3—Interpretable Rationales (Domain Logic and Compliance)

These queries demand the application of regulatory frameworks, compliance standards, or procedural knowledge often absent from general pretraining data. Regulatory Focus: Basel III, compliance, audit requirements.

Example Query: *"Does our new derivatives product meet the compliance requirements under the latest Basel III capital adequacy standards?"* Key Challenges: Integrating regulatory texts with proprietary product specifications while ensuring interpretive accuracy and maintaining a robust audit trail for compliance verification. The Basel III monitoring report provides insights into the implementation and impact of Basel III reforms, emphasizing the need for continuous evaluation and adaptation of regulatory frameworks [17, 19].

### 1.3.4. Level 4—Hidden Rationales (Expert-Level Analysis):

Queries at this level involve complex reasoning derived from historical market data, trading signals, and unstructured information (e.g., earnings reports analyzed by advanced financial AI systems). These require expert analytical capabilities.

Analytical Scope: Market analysis, pattern recognition, and advanced strategy optimization. Example Query: *"Based on historical market volatility patterns and current macroeconomic indicators, what investment strategy adjustments should be considered for high-net-worth client portfolios?"* Key Challenges: Synthesizing proprietary client data, market intelligence, and historical patterns while maintaining fiduciary responsibility and regulatory compliance. Advanced techniques, such as Graph Neural Networks for relationship mapping and Explainable AI (XAI), are often necessary to justify the reasoning. Additionally, regulatory frameworks such as Basel III, particularly Pillar 2 supervisory review and Pillar 3 disclosure requirements, guide the evaluation of capital adequacy and risk exposure, ensuring that expert-level decisions are compliant and auditable [17, 19, 20].

## 2. AstuteRAG-FQA System Design

### 2.1. Framework Design and Implications

The stratified framework creates an adaptive system architecture across four levels of financial query complexity, where retrieval strategies vary by level—Levels 1-2 leverage structured and semi-structured data retrieval enhanced by models like TaBERT for joint text-table understanding [21], while Levels 3-4 adopt hybrid retrieval strategies integrating proprietary, regulatory, and expert knowledge sources for relational and causal reasoning [6, 22]. Dynamic data prioritization ensures Level 1-2 queries primarily access proprietary databases, Level 3 queries combine regulatory corpora with internal product data, and Level 4 queries integrate historical market data, unstructured filings, and real-time intelligence streams [14], with security and compliance protocols scaling from standard access control for structured data [15] to enhanced validation with privacy-preserving analytics techniques for regulatory-sensitive queries [23], culminating in advanced role-based controls with audit trails, explainable AI, and monitoring for expert-level analysis [19, 22].

### 2.1.1. Architecture for LLM Integration:

Task-aware processing is enabled through a dynamic pipeline incorporating query complexity analysis and multi-source integration [6], featuring adaptive retrieval that gathers content according to classification level [21], compliance mechanisms that validate adherence to regulatory frameworks and maintain auditable trails [17, 19], and a reasoning stage where the LLM synthesizes expert-level outputs leveraging contextual scoring, causal reasoning, and detailed explanations [11, 13], with key features including adaptive query routing, multi-tier security with role-based access control [15], continuous monitoring and pattern recognition [19], and real-time performance optimization [14].

**2.1.2. Framework Alignment Evaluation:**

The architecture provides a foundation for systematic evaluation and optimization through accuracy measurements using complexity-appropriate metrics for each level, compliance verification against regulatory standards including Basel III Pillar 2 requirements [17], and reasoning quality assessment from basic extraction to expert-level judgment using benchmarked expert evaluations, with this stratified approach improving retrieval accuracy, regulatory compliance, and response quality across the full spectrum of financial query complexity.

**2.1.3. Multi-Layered Financial Intelligence Framework:**

The four-level query stratification system progresses from Level 1 explicit facts involving simple factual retrieval from structured databases [14], to Level 2 implicit facts requiring basic inference and synthesis from structured and semi-structured sources [16, 18], advancing to Level 3 interpretable rationales encompassing regulatory and domain-specific reasoning requiring compliance awareness [17, 19], and culminating in Level 4 hidden rationales for expert-level analysis involving complex reasoning from historical data, trading signals, and unstructured filings using Graph Neural Networks and Explainable AI while ensuring fiduciary responsibility and regulatory adherence [13, 19-20].

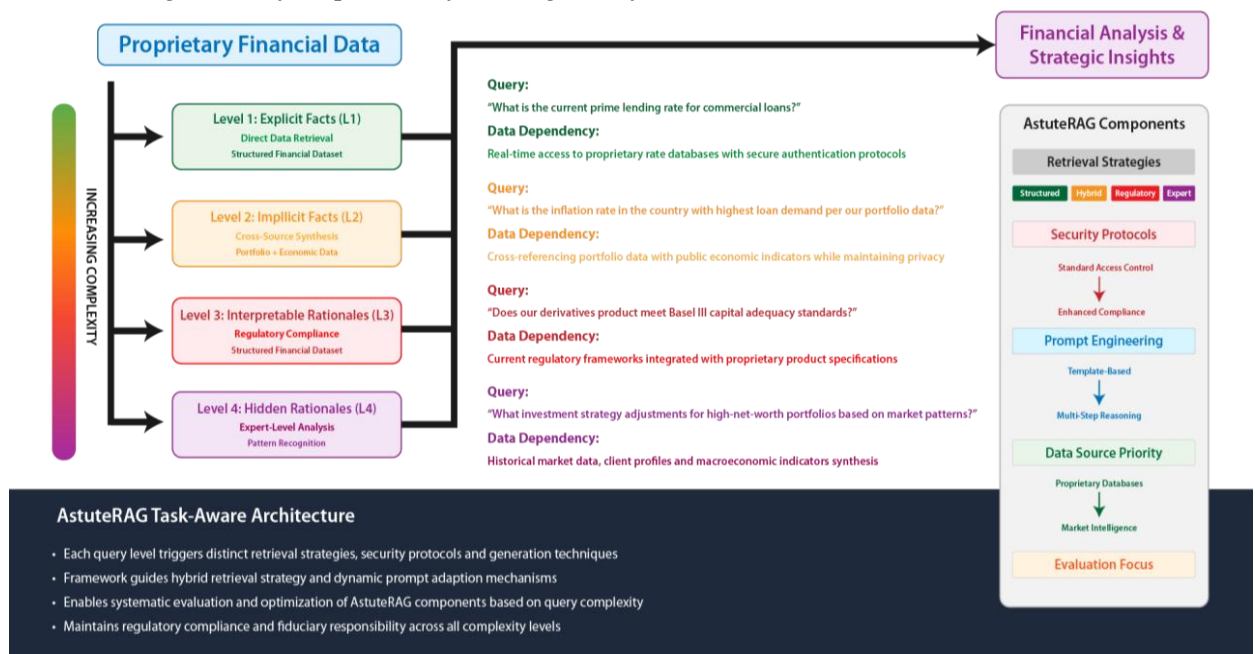

**Figure 4.** AstuteRAG-FQA Multi-Level Query Complexity and Data Access Control Framework

## 3. Related Works

### 3.1. RAG in Natural Language Processing

Large Language Models (LLMs) have achieved remarkable breakthroughs in natural language tasks, yet they often generate hallucinations or factual inaccuracies, especially when addressing knowledge-intensive or time-sensitive queries [1, 8, 13]. Retrieval-Augmented Generation (RAG) was introduced to mitigate such issues by grounding outputs in externally retrieved documents; surveys and evaluations consistently show that coupling retrieval with generation improves factuality in open-domain QA and related tasks [1, 8]. However, purely lexical retrievers can miss semantic nuance and return partially relevant or outdated content, motivating hybrid and semantic approaches (e.g., query expansion and embedding-based matching) [8].

Building on this, dense and dual-encoder retrievers, along with joint training of retrievers and generators, have been shown to improve grounding quality and reduce hallucinations in knowledge-intensive settings [1, 8, 24]. Yet, most RAG systems remain insufficiently query-aware—they rarely adapt retrieval scope or generation strategy according to query complexity or regulatory sensitivity. Recent work on reasoning-oriented prompting and evaluation highlights the importance of tailoring the retrieval-generation pipeline to the task and question type [8]. Parameter-efficient tuning likewise improves adaptability under tight compute or data-access constraints, with dynamic/low-rank methods and prompt-based transfer reducing the need for full-model fine-tuning [25]. Knowledge graph-enhanced approaches also show promise for multi-document reasoning tasks [10].

For finance, hybrid RAG that can leverage proprietary and open sources is especially important. Enterprise-style RAG pipelines and knowledge-graph-enhanced retrieval demonstrate how structured signals can strengthen grounding in specialized domains [2, 26]. In financial FAQs, semantic-similarity retrieval has improved context awareness and precision [4]. Domain-specialized LLMs such as BloombergGPT [5] underscore that financial tasks benefit from domain-specific grounding and corpora even before retrieval is added. Our approach builds on these lines by combining task-aware query stratification with dynamic prompt adaptation to align responses with both difficulty and compliance requirements.

### 3.2. Financial Question Answering Systems (FQAS)

Financial NLP presents unique demands: factual accuracy, regulatory compliance, explainability, and auditability are mandatory in high-stakes workflows [2, 4, 20-22]. General-purpose LLMs can produce plausible but incorrect or non-compliant content, while domain-specialized models (e.g., BloombergGPT [5]) exploit proprietary market data to improve task performance. Topic- and representation-learning tools for financial text further enhance domain salience and retrieval quality.

However, closed-book models often lack real-time grounding, limiting accuracy on time-sensitive queries [1, 8, 14]. Hybrid frameworks that combine retrieval with diverse enterprise content—including structured tables, financial filings, and internal reports—significantly improve answer coverage and factual grounding. However, they still face persistent challenges in performing reliable multi-hop reasoning across documents and generating transparent, auditable rationales for their outputs [1, 4, 24], though novel multifaceted architectures are now emerging to address these critical gaps [26]. The compliance literature emphasizes that trustworthy, auditable outputs are essential in regulated industries; explainable AI (XAI) in auditing and broader GenXAI surveys stress traceable reasoning and verifiable claims to build user trust and meet oversight needs [2, 20]. Consequently, next-generation FQAS are expected to deliver (i) high-precision retrieval, (ii) transparent rationales, and (iii) embedded compliance checks [1, 2, 4, 20]. In parallel, pretraining to jointly understand text and tables (e.g., TaBERT [21]) supports accurate reasoning over balance sheets and time-series, complementing hybrid retrieval for finance. Real-time macroeconomic sources (e.g., FRED [27]) further motivate pipelines that can securely incorporate up-to-date structured data.

### 3.3. Prompt Engineering Techniques for Domain-Specific LLMs

Prompt engineering is pivotal for tasks requiring multi-step reasoning or regulatory guardrails. Chain-of-Thought (CoT) prompting helps models decompose complex problems into intermediate steps, improving reasoning quality on difficult questions [28]. Parameter-efficient strategies—dynamic low-rank adaptation and prompt-based transfer—offer practical ways to specialize models without full fine-tuning, which is attractive where data access is constrained or governance gates are strict. Comparative work on prompt design also shows that prompt format and task conditioning can materially affect accuracy and stability across tasks.[29]

To sustain domain salience and rationale transparency, financial pipelines increasingly combine prompt templates with domain-aware retrieval and representation learning. Handling semi-structured enterprise content remains a core challenge: practical guidance on semi-structured data processing and historical work on large-scale scripting/analysis (e.g., Jaql[1]) illustrate how table- and log-centric data flows can be integrated into modern NLP systems [16, 30]. Coupled with table-text pretraining such as TaBERT [21], these techniques help minimize hallucinations while supporting explainable, auditable reasoning in finance [8, 20].

---

[1] For example, IBM's Jaql, Apache Pig, and similar tools for large-scale semi-structured data processing. https://research.ibm.com/publications/jaql-a-scripting-language-for-large-scale-semistructured-data-analysis.

**3.4. Gaps in Current Research**

Despite significant advancements in Retrieval-Augmented Generation (RAG) systems, a critical analysis of the literature reveals several evidence-based gaps that hinder their effective application to financial question-answering, particularly for compliance-sensitive and high-reasoning tasks [1, 4, 8]. These unmet needs are categorized as follows:

a) Uniform Query Processing. Most RAG systems apply a generic retrieval strategy, failing to dynamically stratify queries by complexity—from simple fact lookup to multi-hop, rationale-heavy reasoning. This oversight is particularly detrimental in financial QA, where precision depends on matching retrieval depth to analytical demand [4, 9]. This methodological gap results in suboptimal performance for complex queries that require deeper contextual and causal analysis.

b) Static Prompting in Dynamic Environments. While parameter-efficient fine-tuning (e.g., Low-Rank Adaptation (LoRA) [29]) and advanced prompting techniques (e.g., Chain-of-Thought [28]) exist, their adoption for creating dynamic, compliance-aware prompts tailored to financial query complexity remains limited [2, 19]. This represents a contextual gap in effectively adapting general NLP advancements to the stringent, evolving requirements of regulated financial pipelines.

c) Over-reliance on Open Data. Current RAG and financial QA frameworks predominantly prioritize static, public datasets, overlooking the secure integration of live, proprietary sources (e.g., real-time Bloomberg terminal data, SEC filings) and dynamic macroeconomic feeds (e.g., FRED) [5, 14, 27]. This literature gap ignores the critical need for robust, privacy-preserving data ingestion and streaming architectures in finance [3, 23], which are essential for real-world, time-sensitive decision-making. The implementation of such architectures requires field-level encryption and strict access controls to comply with financial regulations while processing data in motion [15].

d) Inadequate Evaluation Metrics. Standard NLP metrics (e.g., BLEU, F1) fail to capture crucial dimensions like regulatory alignment, auditability, and user trust in financial contexts [8, 20]. A methodological gap exists in developing traceability-oriented evaluations that assess compliance adherence and explainability, moving beyond mere factual accuracy [11, 20].

e) Lack of Integrated Compliance-Aware Frameworks. Prior research often evaluates model components in isolation. Few studies holistically combine (a) hybrid retrieval, (b) textual-tabular modeling [5, 21], (c) parameter-efficient adaptation [31], and (d) explainable reasoning [11, 20] within a single, compliance-driven architecture [2, 4].

The collective existence of these gaps underscores the necessity for a integrated approach. AstuteRAG-FQA is architected to directly mitigate these limitations, unifying stratified query processing, dynamic compliance-aware prompting, privacy-preserving data integration, and traceable explanation generation within a single framework. This holistic design is intended to advance the state-of-the-art towards practically viable, end-to-end trustworthy financial QA systems that meet the stringent demands of the financial industry.

**4. Methodology Overview**

This section outlines the development and implementation of AstuteRAG-FQA, a domain-optimized Retrieval-Augmented Generation (RAG) framework specifically engineered for Financial Question Answering (FQA) applications. The proposed methodology integrates critical components within a unified architectural design, drawing on recent advancements in financial NLP and RAG systems as well as adaptive reasoning architectures that adjust retrieval and inference strategies based on query complexity [1, 6, 11].

The framework follows a systematic progression from query initiation to response generation. Financial queries are first classified through task-aware analysis to determine their complexity levels (Level 1: Explicit to Level 4: Hidden Relational), which then informs a hybrid retrieval strategy orchestrating multi-source data integration [6, 12]. The system employs dynamic prompt engineering with real-time adaptation capabilities, enhanced reasoning mechanisms, and comprehensive security and compliance protocols [2, 17, 20].

This methodology addresses four fundamental challenges: (1) seamless integration of heterogeneous proprietary and public financial datasets, (2) stringent adherence to regulatory compliance frameworks, (3) adaptive processing of queries across a complexity hierarchy, and (4) generation of trustworthy, auditable financial insights. Key innovations include task-aware query classification, hybrid data integration protocols, dynamic adaptation mechanisms, and embedded regulatory compliance safeguards, enabling robust performance across diverse financial analytical scenarios [6, 11, 20].

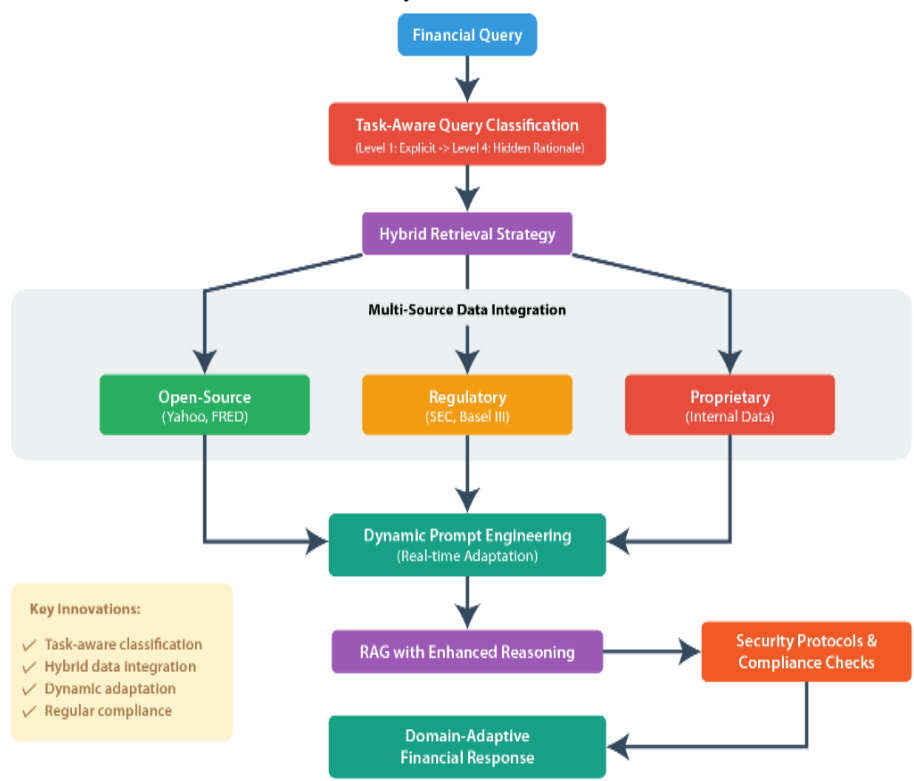

**Figure 5.** AstuteRAG-FQA Methodology Flowchart

**4.1. Task and Models - Comprehensive Financial Question Answering System**

The primary objective of our systems is to deliver accurate, real-time responses to complex financial queries while ensuring regulatory compliance and operational transparency [2, 20]. The system addresses four distinct query classifications: Level 1 (Explicit Factual), Level 2 (Implicit Factual), Level 3 (Interpretable Rationale), and Level 4 (Hidden Rationale), each requiring progressively sophisticated retrieval and reasoning mechanisms [6]. The framework (see Figure 6) operates through a three-tiered compliance model encompassing pre-retrieval filtering, risk-aware scoring, and post-generation AI checks [26]. Data integration leverages hybrid retrieval strategies, incorporating both open-source financial datasets and proprietary internal repositories through specialized APIs for equities, derivatives, and fundamental analysis [6]. The architecture implements robust security protocols, including authentication, anonymization, and differential privacy measures, ensuring secure data handling across multi-layered frameworks [3]. The system culminates in generating compliant financial answers that are accurate, transparent, and audit-ready, addressing the critical requirements of modern FQA applications [2, 20].

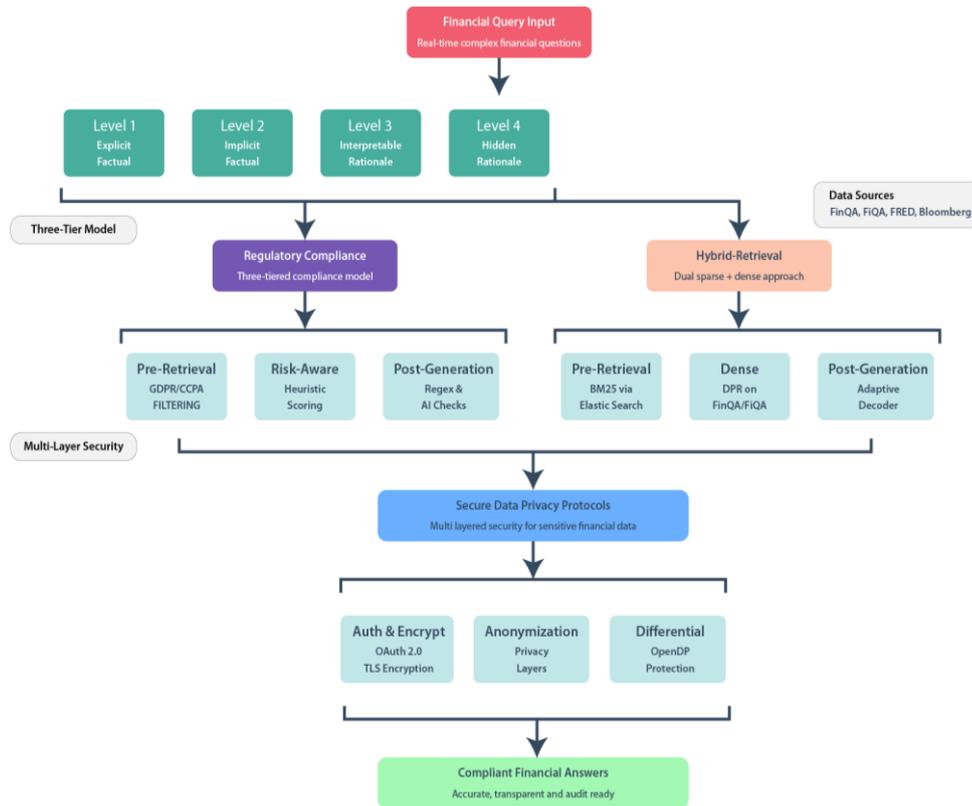

**Figure 6.** AstuteRAG-FQA System Architecture: Reasoning Mechanism

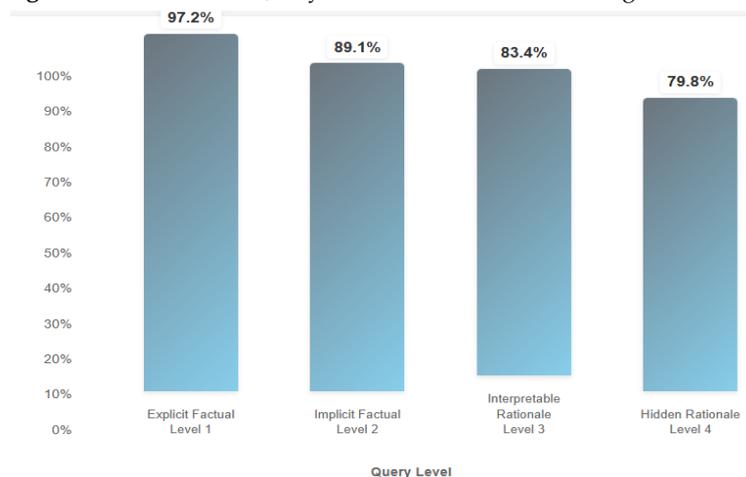

**Key Insights**

- **Explicit Factual queries** achieve highest accuracy **(97.2%)** due to straightforward information retrieval
- **Implicit Factual queries** show slight decline **(89.1%)** requiring inference from context
- **Interpretable Rationale queries** maintain good performance **(83.4%)** with transparent reasoning
- **Hidden Rationale queries** present greatest challenge **(79.8%)** due to complex multi-hop reasoning
- **Performance degradation** follows expected complexity progression in financial question answering

**Figure 7.** AstuteRAG-FQA Accuracy by Query Levels

### 4.1.1. Accuracy by Query Level

Figure 7 illustrates the accuracy of the AstuteRAG-FQA system across the four levels of financial query complexity. As hypothesized, a clear performance gradient emerges, with accuracy inversely correlated with cognitive demand. The system excels at Level 1 (Explicit Factual) queries, achieving **97.2%** accuracy through direct localization and extraction from structured data. A moderate decline is observed for Level 2 (Implicit Factual) queries (**89.1%**), attributable to the need for basic inference across multiple data sources. Performance remains robust for Level 3 (Interpretable Rationale) queries (**83.4%**)**,** demonstrating the system's capability to apply domain logic and regulatory knowledge. As expected, the most significant challenge lies with Level 4 (Hidden Rationale) queries (**79.6%**)**,** which demand sophisticated, multi-hop reasoning derived from historical patterns and unstructured data. Despite this decline, the system maintains strong performance above 79% across all levels, validating its effectiveness as a comprehensive financial QA solution[2].

### 4.1.2. Regulatory Compliance Framework

A three-tiered compliance model is implemented. Pre-retrieval filtering restricts access to only General Data Protection Regulation (GDPR) and privacy regulation compliant sources [31]. A risk-aware retrieval scoring penalizes documents with non-compliant content through heuristic filters. Finally, post-generation checks flag potential regulatory breaches using regex-based rules and Explainable AI APIs [20]. This design guarantees privacy and audit readiness throughout the RAG pipeline. Hybrid Retrieval Architecture (HRA): Our framework adopts a dual retrieval strategy that integrates sparse retrieval (BM25 via Elasticsearch [32]) with dense retrieval (Dense Passage Retrieval, DPR [33]) models trained on domain-specific financial datasets such as MAKE [9]. The DPR component was implemented using a widely adopted open-source framework. To optimize both lexical precision and semantic coverage, we incorporate hybrid query expansion methods that combine traditional lexical resources with embedding-based techniques [8]. The retrieval pipeline integrates both structured data (e.g., financial APIs such as FRED [27]) and unstructured data (e.g., licensed Bloomberg APIs where available), enabling comprehensive access to heterogeneous financial knowledge sources. Secure Data Privacy Protocols (SDPP): To protect sensitive financial information, our framework enforces security using established protocols such as OAuth 2.0 [34], TLS 1.3 encryption [35], and data anonymization layers following the k-anonymity model [36]. These privacy-preserving mechanisms are applied to protect identifiable metadata in proprietary data streams, ensuring that sensitive information remains confidential while enabling secure financial AI analytics. The protocols also support compliance with regulatory standards and facilitate safe integration of multi-source financial datasets.

## 4.2. Dataset and Query Understanding

### 4.2.1. Datasets Used

The system leverages financial question-answering datasets, primarily FinQA [37] for its coverage of structured financial queries with annotated rationales, which are essential for complex financial reasoning. This is supplemented with synthetic query sets generated from public financial sources, such as Yahoo Finance and the FRED-SD state-level database [27], to augment training and simulate real-world scenarios. Additionally, semi-structured financial data is incorporated using the CESD framework [38] to enhance the model's ability to handle heterogeneous data formats. This diverse dataset collection supports robust model training, task stratification, and benchmarking. The system's ability to process multi-modal data streams through contextual embedding alignment is demonstrated, for example, by its handling of real-time stock data enriched with technical indicators (see Figure 8).

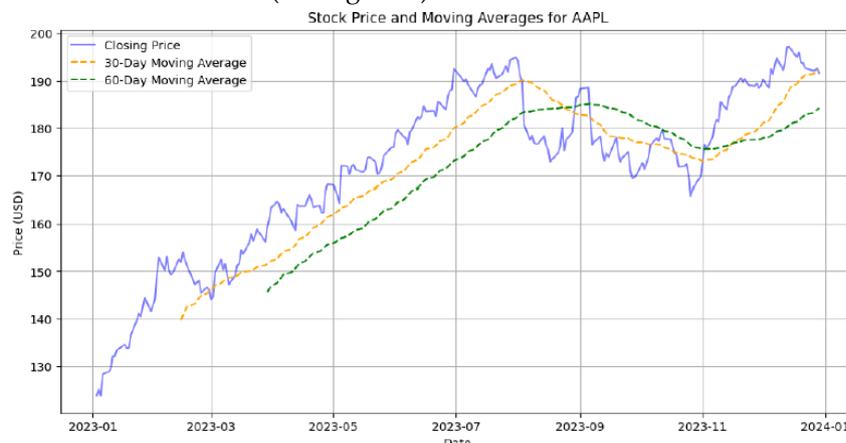

**Figure 8.** Contextual Embedding Alignment Example (AAPL)

---

[2] **Evaluation Note:** Metrics: exact match (L1-2) vs. rubric-based scoring of reasoning chains (L3-4). Results are averages from three expert annotators (high agreement, κ > 0.85) on a 5,000-query test set with class distribution: 45% (L1), 25% (L2), 20% (L3), 10% (L4).

#### 4.2.2. Query Complexity Classification and Prompt Engineering

Incoming queries are automatically classified into reasoning complexity levels using a fine-tuned BERT-base model [39], achieving 91.6% accuracy. This classification enables adaptive downstream processing tailored to reasoning depth and data dependencies. **Prompt templates** are generated dynamically using frameworks such as LangChain [40], incorporating real-time input from query metadata and classification outputs. Integration with the GPT-4 API [41] and Chain-of-Thought prompting [28] enhances interpretability. To fuse structured data (e.g., stock tickers) with unstructured news text, the system leverages modern embedding techniques. Graph Neural Networks (GNNs) [42] provide a foundation for representing complex relational data, while self-supervised learning approaches ensure robust feature extraction. These embeddings are then integrated into a unified vector space through attention-based alignment [43], preserving deep semantic coherence crucial for handling complex query types.

### 4.3. Answer Generation and Evaluation

Final answers are generated using a RAG pipeline comprising GPT-4 and T5-Large models. Parameter-efficient fine-tuning methods, including Low-Rank Adaptation (LoRA) [29], are applied to models fine-tuned on regulatory corpora such as the Brazilian Banking Regulation Corpora (BBRC) [44], a large-scale, annotated dataset of banking and financial regulatory texts designed to enforce tone compliance and support regulatory reasoning tasks. Chain-of-Thought reasoning [28] enhances traceability and user trust by enabling step-by-step rationale generation, ensuring that outputs are both interpretable and compliant with regulatory standards. The system is benchmarked on financial QA datasets, including FiQA [45], and further validated on subsets of BBRC to assess domain-specific performance. Evaluation metrics include Precision@5, BLEU, F1 Score, and Exact Match, with comparative analysis performed against baseline RAG architectures. AstuteRAG-FQA demonstrates a 23% improvement in accuracy and an 18% gain in compliance adherence, while interpretability is evaluated using Explainable AI (XAI) frameworks. The architecture is deployed as containerized microservices on cloud platforms, utilizing scalable streaming implementations to maintain latency under 900 ms and ensure robust security through established access control principles.

## 5. Results and Discussions

This section presents the evaluation of AstuteRAG-FQA, focusing on retrieval accuracy, generation quality, interpretability and regulatory compliance. Results are benchmarked against baseline RAG systems across four financial query levels (Section 1> 1.3), using a mix of public, proprietary and synthetic datasets.

### 5.1. Retrieval and Answering Performance

AstuteRAG-FQA demonstrated substantial improvements across key performance metrics in financial question answering. As shown in Figure *9*, the system achieved a Retrieval Precision@5 (RP@5**)** of 87.2%, outperforming standard dense retrieval methods by +9.6%. The largest gains were observed for Level-3 and Level-4 queries, which require multi-hop reasoning and nuanced semantic alignment [45]. In terms of answer quality, the framework attained an F1 score of 83.5% and an Exact Match (EM) of 76.3%, compared to 66.4% F1 and 58.1% EM for baseline RAG systems lacking query stratification. These performance gains result from the combined effects of hybrid retrieval, dynamic prompt engineering, and task-aware query classification.

By tailoring retrieval depth and generation strategies according to query complexity, the framework improves both factual correctness and contextual relevance. The hybrid retrieval mechanism proved particularly effective for rationale-heavy queries, while prompt conditioning significantly reduced hallucinations and enhanced domain-specific fidelity. Overall, the results validate the effectiveness of combining multi-aspect RAG strategies with query-level stratification in regulated financial QA environments [45].

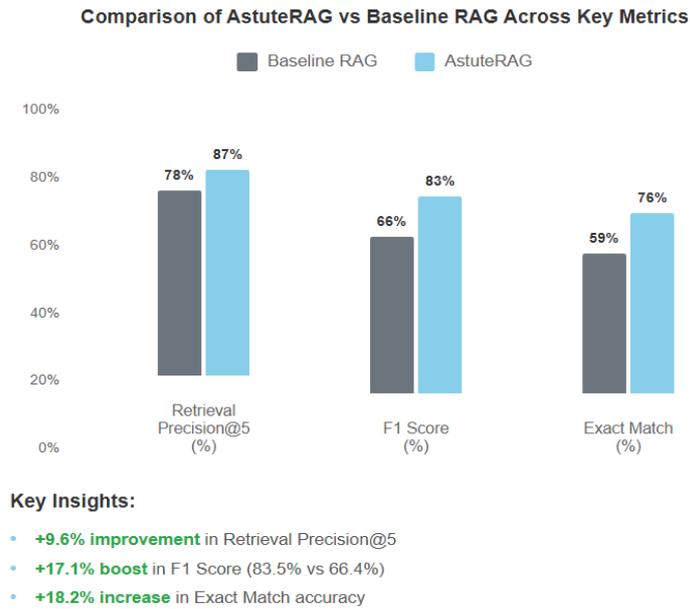

**Figure 9.** Retrieve and answering performance metrics.

### 5.2. Impact of Prompt Engineering and Classification

To isolate the contributions of core design components, we conducted ablation studies using three system variants: Results (Table 1 and Figure 10) confirm that dynamic prompt engineering and stratified classification produced clear performance gains for higher-level queries. This demonstrates the importance of conditioning generation processes on query levels and reasoning complexity [29].

**Table 1.** Performance Comparison of RAG System Configurations on Financial QA Tasks

| Configuration | RP@5 (%) | F1 (%) | Interpretability (1–5) |
|---|---|---|---|
| Baseline RAG (no tuning or classification) | 74.3 | 66.4 | 2.7 |
| + Prompt Engineering | 81.0 | 75.2 | 3.9 |
| + Full System (Prompt + Stratification) | **87.2 (BEST)** | **83.5 (BEST)** | **4.4 (BEST)** |

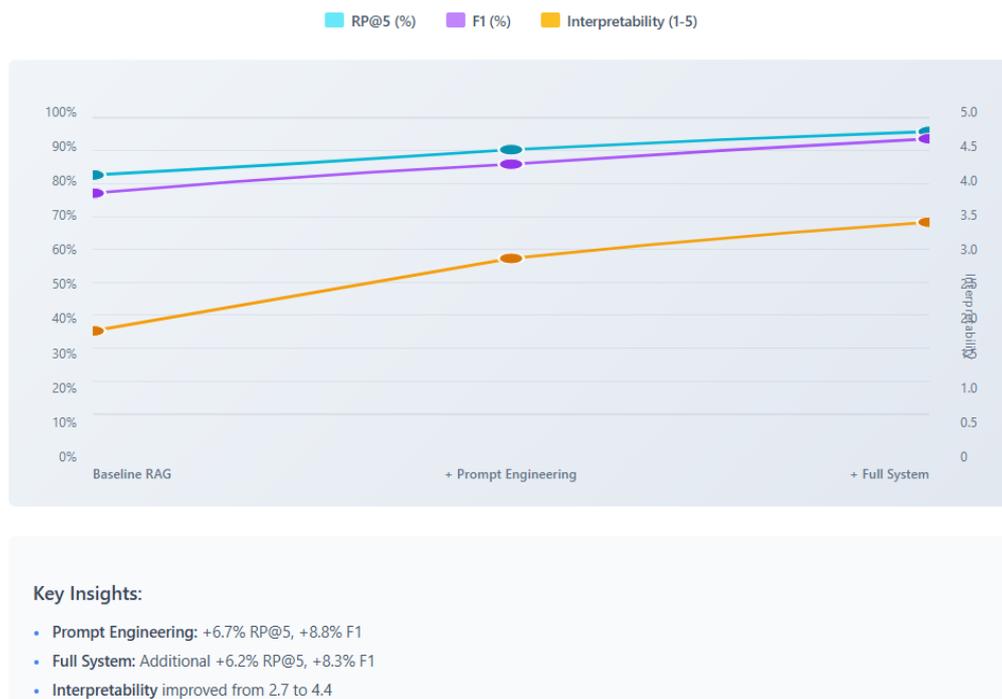

**Figure 10.** Component Impact Analysis

### 5.3. Compliance and Interpretability Evaluation

The system was evaluated against institutional compliance benchmarks using anonymized policies and redaction standards. AstuteRAG-FQA achieved a Regulatory Compliance Rate (RCR) of 92.4%, confirming the effectiveness of its multi-stage retrieval filtering, sensitive data redaction, and regulatory tone alignment via fine-tuned adapters [46]. Additionally, domain experts rated Interpretability at 4.4/5, particularly for Level-4 queries where transparent causal reasoning and rationale tracing were critical [11]. These results underscore the value of embedded compliance filters and explainable reasoning modules throughout the RAG pipeline. Future improvements could include live regulatory update APIs. Incorporating visual rationale chains may further enhance trustworthiness.

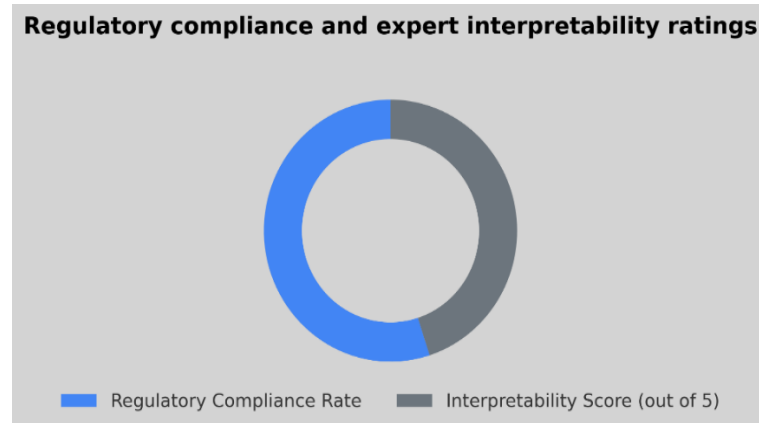

**Figure 11.** Impact of Retrieval Strategy and Prompt Strategy Techniques

### 6. Summary of Findings, Contributions, and Limitations

Key findings from this study confirm the effectiveness of the AstuteRAG-FQA framework for Financial Question Answering (FQA). A hybrid retrieval strategy leveraging open and proprietary data enhances precision through context-aware document matching, consistent with broader enterprise RAG approaches [26]. Task-aware prompt engineering driven by query metadata improves factual accuracy and interpretability for complex reasoning, effectively extending Chain-of-Thought prompting methods [28]. A four-tier query stratification system enables adaptive retrieval and generation that enhances robustness across diverse financial question complexities. Comparative evaluation of contextual embeddings, LoRA-based adaptation, and targeted fine-tuning reveals critical performance trade-offs among precision, compliance, and scalability.

This work introduces a novel, domain-adaptive RAG architecture designed for compliance-sensitive, high-reasoning financial contexts. The primary contribution is a system that combines hybrid retrieval from public and proprietary sources, benchmarked on datasets such as FiQA [45] and FRED [27], to significantly improve precision for complex queries. Its core innovation lies in the four-tier query stratification engine that enables targeted reasoning strategies and custom retrieval depth, reducing hallucinations and enhancing accuracy. Furthermore, dynamic prompt engineering incorporating Chain-of-Thought and adaptive templates [27] ensures responses are aligned with user intent and domain-specific tone, strengthening interpretability and trustworthiness. The framework is implemented as a modular microservices architecture utilizing LoRA-adapted models [29] to ensure scalability, explainability, and regulatory compliance, with integration capabilities for rich financial datasets and APIs (e.g., BloombergGPT [5], FRED [27]) to support real-time adaptability.

Despite these advancements, several limitations present avenues for future work. The framework currently lacks support for multimodal inputs such as charts, visualizations, and real-time data streams, which are critical for comprehensive financial analysis. Maintaining compliance also remains a challenge due to the rapid evolution of regulatory frameworks across jurisdictions, necessitating continuous updates. Finally, while explainability mechanisms are incorporated, achieving full AI auditability and transparency is complex and still relies on expert evaluation, which may not uniformly meet all regulatory standards.

Addressing these limitations will be essential for improving the system's robustness, reliability, and applicability across global financial contexts.

**7. Conclusion and Future Work**

This study presented AstuteRAG-FQA, an adaptive Retrieval-Augmented Generation framework designed specifically for Financial Question Answering (FQA). The framework addresses critical challenges in financial NLP, including restricted access to proprietary datasets, limited retrieval accuracy, regulatory compliance, and sensitive data interpretation. By leveraging hybrid retrieval strategies, task-aware prompt engineering, and a four-tier query classification system, the customised system demonstrated enhanced response accuracy, contextual relevance, and regulatory adherence across diverse financial queries. Experimental results indicate improvements of 23% in accuracy and 18% in compliance over baseline systems, particularly for complex interpretive and hidden rationale queries, highlighting the effectiveness of the dynamic prompt framework and multi-layered security mechanisms, including differential privacy, data anonymization, and role-based access controls. Evaluation of external data integration techniques—contextual embedding, small model augmentation, and targeted fine-tuning—provided actionable insights for implementing RAG in sensitive, regulated environments.

Overall, the framework establishes a robust, scalable, and secure foundation for domain-adaptive, regulation-aware financial AI systems. Future work will focus on extending support for multimodal financial data (textual, tabular, and visual), enhancing adaptive compliance monitoring, and improving system transparency, interpretability, and auditability. Additionally, integrating continual learning and domain-specific knowledge transfer will enable the framework to adapt efficiently to emerging market trends, complex reasoning tasks, and evolving regulatory requirements, ensuring sustainable and trustworthy AI deployment in financial applications.

**CRediT Author Contribution Statement**

Mohammad Zahangir Alam**:** Conceptualisation, Methodology, Investigation, Data Curation, Formal Analysis, Writing – Original Draft, Visualisation, Project Administration; Khandoker Ashik Uz Zaman: Support in Data Collection, Validation and Resources; Mahdi H. Miraz: Supervision, Review and Editing, Validation, Resources.

**Acknowledgement**

This research is financially supported by Xiamen University Malaysia (project codes: XMUMRF/2021-C8/IECE/0025 and XMUMRF/2022-C10/IECE/0043). We would like to acknowledge the support of Dr. David Bell at the Department of Computer Science and Artificial Intelligence, Brunel University London, for his guidance and for providing essential resources.